\documentclass{article}
\usepackage{ijcai19}

\usepackage{times}
\usepackage{soul}
\usepackage{url}

\usepackage[hidelinks]{hyperref}
\usepackage[utf8]{inputenc}
\usepackage[small]{caption}
\usepackage{color}
\usepackage{graphicx}
\usepackage{amsmath}
\usepackage{booktabs}
\usepackage{algorithm}
\usepackage{algorithmic}
\usepackage{booktabs} 
\usepackage{amsthm} 
\usepackage{multirow}
\usepackage{gb4e}
\noautomath 
\title{Early Discovery of Emerging Entities in Microblogs}

\author{
    Anonymous Authors
    \affiliations
    \emails
}

\author{
Satoshi Akasaki$^1$\and
Naoki Yoshinaga$^2$\And
Masashi Toyoda$^2$\\
\affiliations
$^1$The University of Tokyo\\
$^2$Institute of Industrial Science, the University of Tokyo\\
\emails
\{akasaki, toyoda\}@tkl.iis.u-tokyo.ac.jp,  ynaga@iis.u-tokyo.ac.jp
}

\newtheorem*{dfn:ec}{\textbf{Emerging contexts}}
\newtheorem*{dfn:pc}{\textbf{Prevalent contexts}}
\newtheorem*{dfn:ee}{\textbf{Emerging entities}}
\newtheorem*{dfn:pe}{\textbf{Prevalent entities}}
\newtheorem*{dfn:le}{\textbf{Long-tail entities}}
\newtheorem*{dfn:he}{\textbf{Homographic entities}}

\hypersetup{final}
\begin{document}
\maketitle

\begin{abstract}
Keeping up to date on emerging entities that appear every day is indispensable for various applications, such as social-trend analysis and marketing research. Previous studies have attempted to detect unseen entities that are not registered in a particular knowledge base as emerging entities and consequently find non-emerging entities since the absence of entities in knowledge bases does not guarantee their emergence. We therefore introduce a novel task of discovering truly emerging entities when they have just been introduced to the public through microblogs and propose an effective method based on time-sensitive distant supervision, which exploits distinctive early-stage contexts of emerging entities. 
Experimental results with a large-scale Twitter archive show that the proposed method achieves 83.2\% precision of the top 500 discovered emerging entities, which outperforms baselines based on unseen entity recognition with burst detection.
Besides notable emerging entities, our method can discover massive long-tail and homographic emerging entities.
An evaluation of relative recall shows that the method detects 80.4\% emerging entities newly registered in Wikipedia; 92.4\% of them are discovered earlier than their registration in Wikipedia, and the average lead-time is more than one year (571 days).
\end{abstract}

\section{Introduction}
\label{sec:intro}
Understanding the latest world events is an important objective for many applications such as social-trend analysis, marketing research, and reputation management. Such applications often require knowledge of \textit{emerging entities}, such as new products, works, and individuals, which ceaselessly emerge one after another, for real-time monitoring of their activities. 
For example, social listening companies such as Salesforce and Oracle that monitor customer reputations need to track new product trends including those of customer's competitors.
Although knowledge bases (KBs), such as Wikipedia, could be used as a reference list of entities, there is a certain delay until the emerging entities are registered in those KBs, and only notable entities are selected for registration. Therefore, instead of relying on KBs, we need to discover as many emerging entities as possible including \textit{long-tail (less frequent but wide variety of) emerging entities} that are mostly overlooked in KBs before they become prevalent or their information appears frequently.

Previous studies~\cite{nakashole2013,hoffart2014,wu2016,farber2016} focus on detecting out-of-KB entities which are not registered in a particular KB, and consequently find massive non-emerging entities, since the absence of the entities in KBs does not guarantee their emergence (\S~\ref{sec:related}). Extracting emerging entities from the obtained out-of-KB entities is difficult since out-of-KB entities are mostly mere long-tail entities and we cannot expect many contexts to judge their emergence. The contexts can be even noisy when they are \textit{homographic emerging entities} (\textit{e.g.,} Go for new programming language and classical board game). Even worse, it is problematic to prepare training (and evaluation) datasets for out-of-KB detection since we need to manually annotate entities that are not registered (but notable) on the basis of the specific state of the given KB.


\begin{table}[t]
\centering
\small

\begin{tabular}{p{0.95\linewidth}}
\toprule
Nintendo \textit{announces} \textbf{Super Smash Bros. Ultimate} - the game features every single  
character from past games. 
\textit{Release data: December 7, 2018} \\
\midrule
JohnnyDepp is set to play celebrated war photographer W. Eugene Smith in \textit{upcoming drama} ``\textbf{Minamata}'' which HanWay Films \textit{will launch at the upcoming} AFM. 
\textit{Filming starts in Japan then Serbia in January 2019} \\
\midrule
\textit{Can't wait} for this one. \textbf{Big Dom's Bagel Shop} \textit{will open} Aug. 25 in Cary. Here are all the details on Pizzeria Faulisi getting into the bagel business 
URL \\
\bottomrule
\end{tabular}
\caption{Example tweets on emerging entities (bold) with expressions suggesting their emergence (italic)}
\label{table:emergingentities}
\end{table}


Considering these difficulties, we introduce a novel task of discovering emerging entities in a microblog when they have just introduced to the public through the microblog (\S~\ref{sec:definition}). This task is more solid than the existing out-of-KB entity classification task since the task definition is independent of a particular KB. 
In our task, we use the fact that people write about emerging entities with \textit{expressions suggesting their emergence} when those entities are not well known to the public (Table~\ref{table:emergingentities}), considering that potential readers would be unfamiliar with them. By taking advantage of these contexts, we can effectively discriminate emerging entities from prevalent ones, even if they are long-tail or homographic emerging entities, and can find them in the early-stage of their appearance.

To obtain contexts of emerging entities, we propose a time-sensitive distant supervision method based on distant supervision~\cite{mintz2009}.
Our method collects early-stage posts in a massive amount of time-series text where non-homographic entities registered in a KB first emerge. At this time, we also collect adequately-later posts after the first appearance as negative examples to robustly discriminate them from emerging contexts. We then train sequence-labeling models from those contexts to discover emerging entities.

We applied our method to our large-scale Twitter archive and compared the discovered emerging entities with those obtained with  baselines, which regards entities that are unseen in a KB~\cite{nakashole2013} or our Twitter archive as emerging. Experimental results showed that the proposed method effectively detected emerging entities in terms of precision of the acquired entities including homographic and long-tail emerging entities. As the evaluation of relative recall and detection immediacy, using the entities newly registered in Wikipedia as a reference, our method detected most entities in the reference, and in most cases, these entities were discovered earlier than their registration in Wikipedia.

Our contributions are as follows: 
\begin{itemize}
\item We introduce a novel task of discovering emerging entities in microblogs as early as possible.
\item We propose a time-sensitive distant supervision method for easily and automatically constructing a large-scale training dataset from microblogs. 
\item Our method 
found emerging entities
accurately (high precision), abundantly (high recall), and quickly (substantially earlier than their registration in Wikipedia).
\item We will release all the datasets (tweet IDs)\footnote{\url{http://www.tkl.iis.u-tokyo.ac.jp/~akasaki/ijcai-19/}} used in experiments to promote the reproducibility.
\end{itemize}

\section{Definition of Emerging Entity}
\label{sec:definition}
In this section, we define what is meant by the term \textit{emerging entity} in this study. 
Our definition of emerging entity is motivated from the report of \cite{graus2018} and meets requirements for social-analysis applications.

\citeauthor{graus2018}\ analyzed how newly registered entities in Wikipedia have appeared in news and social media before they are registered as individual articles. They found that most of those entities shift from the state of ``sporadically mentioned in news and social media'' to that of ``established as one article due to enhancement of references.''



Fortunately, when users submit posts about entities that appeared newly but are not famous yet to social media, 
they usually indicate the emergence of the entities, as in Table~\ref{table:emergingentities}, despite their popularity. We thereby define emerging entities in terms of how they are described in contexts, in other words, how their state is perceived by people as follows:
\begin{dfn:ec}
Contexts in which the writers assumed the readers do not know the existence of the entities. 
\end{dfn:ec}
\begin{dfn:ee}
Entities in the state of being still observed in emerging contexts.
\end{dfn:ee}
We later confirm the solidness of these definitions by evaluating inter-rater agreement of emerging entities acquired from text (\S~\ref{subsec:results}). 
We also define other terms on entities as follows: 
\begin{dfn:pc}
Contexts in which the writers assumed the readers know the existence of the entities. 
\end{dfn:pc}
\begin{dfn:pe}
Entities in the state of being mainly observed in prevalent contexts.
\end{dfn:pe}
\begin{dfn:le}
Entities that are less frequent individually but have wide varieties.
\end{dfn:le} 
\begin{dfn:he}
Entities that share the namings with other entities. 
\end{dfn:he} 

\section{Related Work}
\label{sec:related}
To the best of our knowledge, there has been no study attempting to find emerging entities in microblogs. We review the current tasks related to our task and clarify the term ``emerging entities,'' which has various meanings.

\subsubsection{Emerging and Rare Entity Recognition}
This is a task organized at the 2017 Workshop of Noisy User-generated Text (WNUT 2017)~\cite{derczynski2017} and focused on recognizing both ``emerging and rare'' entities from text. With this task, named entities (NEs) that appeared zero times in a specific (past) portions of datasets are regarded as emerging entities, and manually annotated NE tags to these entities as the target of detection regardless of the contexts in which they have appeared. The dataset used in this task includes the following example (the target entities are in bold):
\begin{quote}
... found photo storage tank that is 5x size of my \textbf{iPhone} with less capacity than \textbf{iPhone 4} ... 
\end{quote}
Consequently, this task is designed to detect (past) data-dependent emerging entities even after they become known to the public (\textit{e.g.,} iPhone). The definition of emerging entities based on a specific data makes it difficult to distinguish emerging entities from prevalent entities. In fact, the 
state-of-the-art
model achieved an F$_1$ of 
$49.59\%$,
which is much lower than usual named-entity recognition (NER) 
on a dataset such as CoNLL-2003 
(F$_1$ of $93.18\%$)~\cite{akbik:2019}. 

Our task discovers emerging entities when they are introduced in microblogs. This 
enables us to take advantage of the fact that emerging entities tend to show their emergence at the early-stage of their appearance. 

\subsubsection{Out-of-KB Entity Identification on News Articles}
This task has been studied to identify NEs that are not registered in a KB (referred to as ``emerging'' entities in the following studies but as out-of-KB entities here for clarity). Since this task is intended to detect entities absent in the KB, it does not distinguish emerging entities from mere long-tail entities.

\cite{nakashole2013} proposed a method for extracting NEs using NER and regards all extracted NEs as out-of-KB if they are not registered in a KB. Since this method ignores contexts in which NEs appear, if the target NE has homographic entities in the KB, it is wrongly classified as an in-KB entity regardless of its emergence (false negatives). Similarly, if the target NE appears with unseen surface (mention), it is wrongly classified as an out-of-KB entity (false positives).

\cite{hoffart2014}, \cite{wu2016}, and \cite{farber2016} proposed methods of classifying whether a given NE in a news article is out-of-KB. Their task is part of the task solved  by~\cite{nakashole2013} since the target NEs are given (assumed to be recognized). Note that NEs are, however, not easily recognizable for languages in which NEs are not capitalized (\textit{e.g.,} German, Chinese, and Japanese). In addition, their methods do not scale to ever-increasing emerging entities because the manual annotations of out-of-KB entities depend on the specific state of the KB and the approaches (and features of the classifier) are tailored for news text.

In contrast to these studies, we focus on ``truly'' emerging entities that are defined independent of KBs and develop an early-detection method of them using a dataset constructed by time-sensitive distant supervision. We targeted microblogs, \textit{i.e.,} timely social media, as sources for emerging entities since \cite{graus2018} reported that emerging entities appear on social media more and earlier than in news articles.

\subsubsection{Notable Account Prediction on Twitter}
This task is to discover long-tail ``rising'' entities (\textit{e.g.,} rising brands) that are expected to be notable in the future within Twitter~\cite{brambilla2017}. Although this task uses Twitter as the source of entities, the same as ours, it requires experts to provide example notable entities. Also, since the target entities are limited to only those with Twitter accounts, it cannot acquire various types of entities that are not linked to Twitter accounts.
We also focus on Twitter but discover emerging entities (\S~\ref{sec:definition}) without relying on domain experts and without restricting the types of entities to be discovered. 


Overall, these related studies define labels (emergence or rare, out-of-KB, or notability) based on specific past data, KBs, or domain experts and annotated them manually. 
We compare our method with two baselines that detect unseen NEs in a KB~\cite{nakashole2013} or in the past Twitter (the same setting as WNUT17) as emerging. We chose these methods because they are only methods applicable to our task, which do not rely on manually annotated data.

\section{Proposed Method}
\label{sec:proposal}
The proposed method discovers emerging entities in microblogs. We target a microblog (Twitter) since \cite{graus2018} reported that compared to news articles, a more diverse range of emerging entities appear earlier on social media, and generally speaking, microblogs include the most timely posts among various types of social media. Note that we do not exploit Twitter-specific functions with our method; thus, it is also applicable to other microblogs such as Weibo. 

To build a supervised model for discovering emerging entities, we exploit the fact that emerging entities are likely to appear in specific contexts (\S~\ref{sec:definition}). By properly identifying such contexts, we can discover corresponding emerging entities effectively and instantaneously even if they are long-tail or homographic ones. The major challenge lies in how to collect such emerging contexts as the training data. 
To cover various emerging contexts for a diverse range of entity types, we develop a method that automatically collects such contexts and corresponding emerging entities.

\subsection{Time-sensitive Distant-supervision}
\label{subsec:distant_supervision}
To meet the expected requirements on the training data for this task, we developed the proposed method (Figure~\ref{ds}) based on time-series text and the distant supervision~\cite{mintz2009}, which automatically collects training data using an existing KB for a specific knowledge-acquisition task. Since our method does not incur any annotation cost, it is easy to prepare and construct the training data. The major difference from the original distant supervision is that labels are not defined only with the KB. We utilize the nature of time-series text to obtain labels for training a emerging entity recognizer.

The idea is to first extract non-homographic entities with unique namings from a KB that emerge when microblog posts are available and to collect their emerging contexts from the time-series microblog posts. The procedure is as follows:

\begin{figure}[t]
    \centering
    \includegraphics[width=0.52\textwidth]{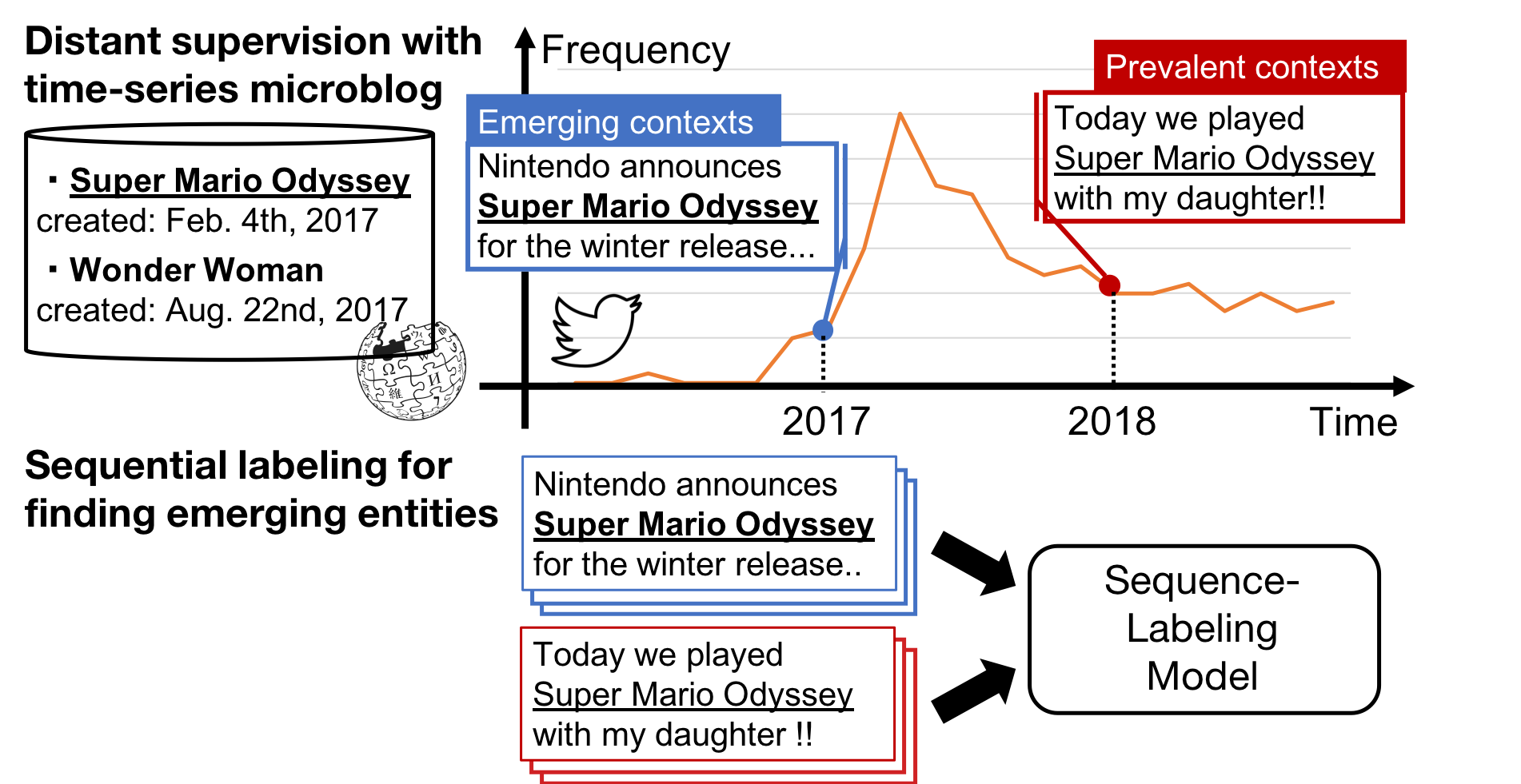}
    \caption{Time-sensitive distant supervision: for the entities retrieved from a KB, emerging and prevalent contexts are collected from microblogs, and sequence labeling models are trained from the obtained emerging and prevalent contexts}
    \label{ds}
\end{figure}

\subsubsection{Step 1 (Collecting candidates of emerging entities)}

We start by collecting titles of articles in Wikipedia as existing entities and then associate them with the time-stamps of registration to collect emerging entities that newly appeared within the available period of the microblog (Twitter). We exclude entities that appeared on Twitter more than $k$ times in the first one-year period where microblog posts are available. This is to exclude homographic entities that share the naming with prevalent entities since it is difficult to collect their emerging contexts only by searching the entities. 

\subsubsection{Step 2 (Collecting contexts of emerging entities)}
For each entity obtained in Step 1, we then retrieve first $n$ early-stage microblog posts posted before the time-stamps of registration as emerging contexts.
Although contexts of long-tail emerging entities are not covered in the obtained training data, similar emerging contexts can be shared by other entities in the KB\@. This is because if the coarse type of entities are the same, their emerging contexts tend to be common regardless of their popularity (\textit{e.g.,} product types tend to be introduced with the term \textit{released}). 

There are two issues to be addressed: 1) how to filter noisy examples of emerging contexts and 2) how to prevent overfitting that detects only the entities used for collecting training data. We explain how we address these issues.

\subsubsection{Filtering Noisy Emerging Contexts}
Although distant supervision can generate abundant training data, incorrectly labeled data can also be included. We therefore collect only reposts (retweets) from the day when the included entities first appeared in retweets more than $k'$ times. This is inspired from the report of~\cite{graus2018} that emerging contexts are likely to be shared by many users since they include information novel to the public. 

\subsubsection{Collecting Prevalent Contexts as Negative Examples}
When a model is trained only with the collected emerging contexts, it will be overfitted to detect only mentions of the emerging entities used to collect the training data. To avoid this, in Step~2, we collect prevalent contexts for the same entities collected in Step~1 as negative examples (Figure~\ref{ds}). Specifically, as the prevalent contexts for each entity, we collect the same number of microblog posts one year after the time of collecting emerging contexts. This enables the model to discriminate between emerging contexts and prevalent contexts and reduces the effect of noisy (prevalent) contexts incorrectly included in the positive examples.

We finally label only the acquired entities in the emerging contexts as emerging entities and combine them with their prevalent contexts to form the training data for sequence labeling
described below. 
We tried several values for the three hyperparameters of our method, $k$, $n$, and $k'$, and confirmed that the accuracy of the models trained from the resulting training data did not markedly change. We therefore empirically set the parameters to $k=5$, $n=100$ and $k'=10$.


\subsection{Sequence Labeling for Finding Emerging Entities}
\label{subsec:sequence_labling}
We next train a sequence-labeling model for finding emerging entities from the collected training data.\ We adopted and compared classical conditional random field (CRF)~\cite{lafferty2001} and modern long short-term memory (LSTM) with CRF output layer (LSTM-CRF)~\cite{lample2016} as the sequence-labeling models. We adopted BIOES as the tagging scheme, which was reported to be better than other schemes \cite{ratinov2009}. We tagged emerging entities in positive examples with BIES and the others with O.

Because our task includes detecting NEs, we referred to features for NER to solve our task. As the CRF features, we use part-of-speech tags, character types, and the results of NER\footnote{We used CaboCha (\url{https://taku910.github.io/cabocha/}).} for the posts, and cluster IDs~\cite{miller2004} obtained from Brown clustering~\cite{brown1992} for each token in input and the two tokens before and after that. LSTM-CRF inputs a word embedding and character embeddings encoded by character LSTM of each token into bi-directional LSTM, which are followed by the CRF layer.

\section{Experiments}
\label{sec:experiment}
We applied the proposed method to actual Twitter archive and performed our task of discovering emerging entities. 

\subsection{Data}
\label{subsec:data}
We constructed the training dataset by using the time-sensitive distant supervision detailed in \S~\ref{subsec:distant_supervision} from our Twitter archive, which we have been compiling since March 11th, 2011 (more than 50 billion tweets have been accumulated).

\begin{table*}[!t]
\footnotesize
    \centering
        \begin{tabular}{@{\,}l@{\quad}l@{\,\,\,}r@{\,\,\,}r@{\quad}l@{\,}}
            \toprule
            \multicolumn{2}{@{\,}l@{\,\,\,}}{\textbf{TYPE}} & \textbf{\# entities} & \textbf{\# posts} & \textbf{examples of emerging context (translated and truncated)}\\
            &\textbf{DBpedia types}\\
            \midrule
\multicolumn{2}{@{\,}l@{\,\,\,}}{\textsc{\textbf{Person}}} & \textbf{4932} & \textbf{23939}\\
& Actor & 885 & 2863 & ... who is the partner of me has changed her name from Mika Tadokoro to \textbf{Momo Nonomiya}\\ 
& MusicalArtist & 731 & 4616 & An idol unit who can do fishing, ``\textbf{TSURIxBIT}'' debuts on May 22th!\\
& SoccerPlayer & 531 & 2327 & We are pleased to announce that \textbf{Kiyotaka Miyoshi} has joined Shimizu S-Pulse.\\
& VoiceActor & 484 & 1596 & Expected new voice actor ``\textbf{Sora Amamiya}'' appeared for the first time on  live broadcasting! ...\\
& BaseballPlayer & 419 & 4390 & ... announced that they have agreed to sign a player contract with \textbf{Spencer Patton}. \\
& AdultActor & 299 & 1731 & ... \textbf{Iori Furukawa}, who debuted last month is also cute! SOD's newcomer is always amazing!!\\
& Model & 281 & 1343 & Nice to meet you, I am \textbf{Tamotsu Kansyuji} from the Juno-Super-Boy Contest. From now on ...\\
& Politician & 260 & 1164 & In Kawasaki Mayor's election, Mr. \textbf{Norihiko Fukuda} defeated other candidates and ... \\
& Person & 177 & 729 & Former CIA official \textbf{Edward Snowden} has revealed the US intelligence gathering\\
& Writer & 152 & 542 & Ms. \textbf{Daruma Matsuura}, who won the newcomer award seems to start serializing from ... \\
& Others (19 types) & 713 & 2638 & You'd better follow the youngster, \textbf{Atsugiri Jason}, who has been working for 2 months ... \\
\midrule
\multicolumn{2}{@{\,}l@{\,\,\,}}{\textsc{\textbf{Creative Work}}} & \textbf{6460} & \textbf{47267}\\
& MusicSingle & 1321 & 11685 & The title of Nana Fujita's single has been decided as ``\textbf{Right Foot Evidence}''! \\
& TelevisionShow & 1153 & 8478 & [Kayoko Okubo] TBS's new program ``\textbf{o-ku bon bon}'' start from today! 24:50-25:20 on air \\
& MusicAlbum & 970 & 6092 & Kis-My-Ft2 3rd album ``\textbf{Kis-My-Journey}'' (provisional) will be released on July 2 this summer! \\
& Film & 917 & 6307 & ... announced that the title of the rebooted version Spider-Man``\textbf{Spider-Man: Homecoming}''.\\
& VideoGame & 652 & 6355 & Latest videos and key art of ``\textbf{DARK SOULS III}'' released! [E3 2015] \\
& Manga & 623 & 2561 & It was announced on Shonen Jump released today, new series ``\textbf{My Hero Accademia}'' starts ... \\
& Anime & 323 & 2983 & Kyoto Animation's TV anime ``\textbf{Tamako Market}'' started broadcasting in January 2013!  \\
& RadioProgram & 266 & 1010 & ... as we will record the first broadcast \textbf{Sasara Night of Fujioka Minami} on the STV. \\
& Book & 146 & 983 & Congratulations on \textbf{Re: Zero-Starting Life in Another World} for upcoming publication! \\
& Others (5 types) & 89 & 813 & KADOKAWA and Hatena's novel posting site \textbf{Kakuyomu} will open on February 29, 2016. ...\\
\midrule
\multicolumn{2}{@{\,}l@{\,\,\,}}{\textsc{\textbf{Location}}}&\textbf{371}&\textbf{1554}\\
& Building & 121 & 756 & A new gourmet building ``\textbf{Ueno no Mori Cherry Terrace}'' is born in Ueno. Both lunch and ... \\
& Museum & 42 & 184 & \textbf{Kumagai Morikazu Tsukechi Museum of Art} opening ceremony commenced. Director ...\\
& Station & 34 & 115 & ... the name of the station to be built at the JR Nambu branch line is decided as \textbf{Odaei Station}!\\
& Settlement & 28 & 47 & We started construction of a new town in \textbf{Slavticci}, 50 km west of the Chernobyl power plant. \\
& School & 24 & 34 & I attended the \textbf{Kaishi International High School} Opening Ceremony. \\
& City & 17 & 46 & Currently, I am in the core city of \textbf{La Hadadatu}, about 130 kilometers away from the war area ...\\
& University & 14 & 47 & I was scheduled to attend the symposium on the establishment of \textbf{Akita University of Art} ...\\
& Others (18 types) & 91 & 325 & Although it is late at night, we have released the \textbf{Ogijima Library} website!\\
\midrule
\multicolumn{2}{@{\,}l@{\,\,\,}}{\textsc{\textbf{Group}}}&\textbf{366}&\textbf{2173}\\
& Company & 259 & 1441 & With the entry of robot business, Softbank established the new company named \textbf{Cocoro} ...\\
& SoccerClub & 55 & 304 & Via Tin Kuwana changed team name to ``\textbf{Vir Tin Mie}'' Press release is available! \\
& Organization & 28 & 179 & Mr. Ishiba decided on Friday to set up a political faction and name it ``\textbf{Suigetsukai}.''  \\
& PoliticalParty & 24 & 249 & ... considers a new political party = Breaking up from nuclear power ``\textbf{Japan's Future Party}'' \\
\midrule
\multicolumn{2}{@{\,}l@{\,\,\,}}{\textsc{\textbf{Other}}}&\textbf{130}&\textbf{561} \\
& Species & 77 & 337 & New species called \textbf{Sado flog}, found on Sado Island, features yellow feet and yellow belly. \\
& CelestialBody & 17 & 75 & [With image] A new Earth-like planet ``\textbf{Grisee 832c}'' is discovered\\
& Others (8 types) & 36 & 149 & Hitachi unveils new ``\textbf{Class 800 Series}'' for high-speed railways in the UK. \\

\midrule
\multicolumn{2}{@{\,}l@{\,\,\,}}{\textsc{Unmapped}} & 7345 & 35552 & JMA named the recent heavy rain as ``\textbf{Heavy Rainfall in  } / \textbf{Kanto Tohoku H27.9}''.\\
&&&&DoCoMo's summer new model ``AQUOS PHONE ZETA \textbf{SH-09D}'' quick photo review ...\\
         \midrule
\multicolumn{2}{@{\,}l@{\,\,\,}}{\textbf{Total}} & 19604 & 111046\\
			\bottomrule
			
        \end{tabular}
\caption{Statistics of the emerging entities and their contexts 
obtained from our Twitter archive by our time-sensitive distant supervision}\label{table:ds}
\end{table*}

\begin{figure*}[htbp]
    \centering
    \includegraphics[width=1.0\linewidth]{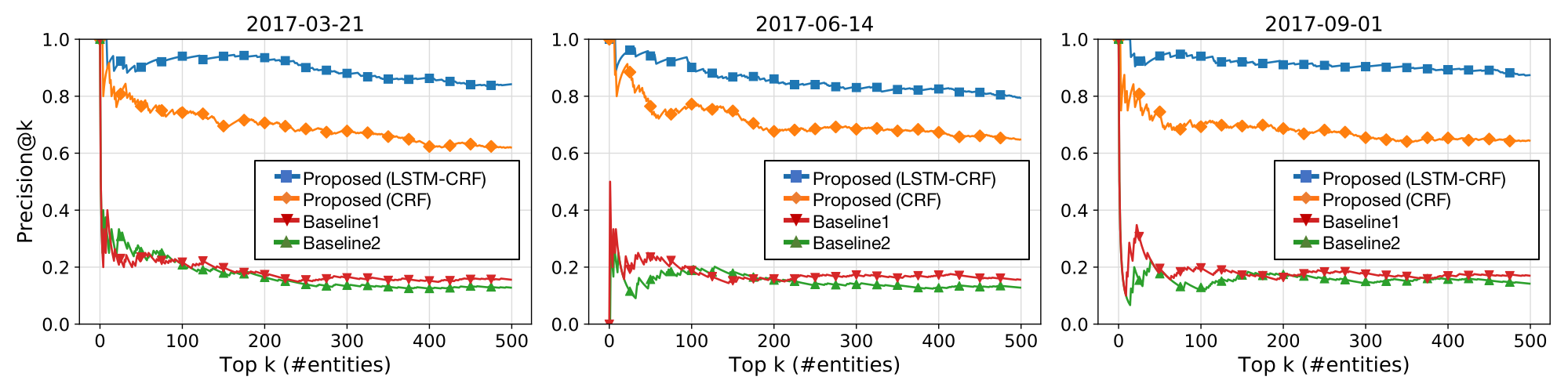}
    \caption{Precision@k for the top-500 emerging entities obtained from Twitter streams by each model.}
    \label{pr}
\end{figure*}

In Step 1 of \S~\ref{subsec:distant_supervision}, we collected titles of articles that were registered in the Japanese version of Wikipedia from March 11th, 2012 to December 31st, 2015 using the Wikipedia dump on June 20th, 2018. We then excluded redirects and disambiguation pages from the titles, and then ran Step 2. We obtained a total of 222,092 tweets including the same number of emerging and prevalent contexts for 19,604 entities as the training data. For model selection, we used 10\% of the training data as the development data. We tokenized each example by using MeCab~(ver.~0.996)\footnote{\url{https://taku910.github.io/mecab/}} with ipadic dictionary (ver. 2.7.0) and then removed URLs, usernames, and hashtags.

We then analyzed the obtained emerging contexts by mapping the included emerging entities to their corresponding types assigned in the DBpedia ontology; for example, the entity ``Spider-Man: Homecoming'' is mapped to the type ``Film.'' Out of the 19,604 emerging entities in our dataset, we have 12,259 type-mappings (51 types). As shown in Table~\ref{table:ds}, the entity types that are manually categorized into \textsc{Person} and \textsc{Creative Work} account for a large proportion. This is because these entities tend to generate a great deal of attention at the time of their appearance than other entities. The unmapped entities included artifacts (\textit{e.g.}, devices, products), Web services, and other terminology because there are no mappings for them in the DBpedia ontology. 
We also see that emerging contexts could be diverse according to the type of entity they include. We thus have to capture those contexts properly to discover various types of emerging entities.

\subsection{Models}
The following models were implemented for comparison:

\paragraph{Proposed (CRF).}
We used the implementation using MALLET~(ver.~2.0.6)~\cite{McCallumMALLET} with L-BFGS as an optimizer. The hyperparameter C was tuned to 0.125 using the development data. 
To obtain Brown clusters, we applied Brown clustering with 1024 clusters to 200 million Japanese tweets sampled from March 11th, 2011 to March 11th, 2012.

\paragraph{Proposed (LSTM-CRF).}
We used the implementation using Theano~(ver.~0.9.0) provided by~\cite{lample2016}.\footnote{\url{https://github.com/glample/tagger/}} We set hyperparameters as suggested in~\cite{yang2018}, who explored the practical settings of neural-sequence-labeling. We optimized the model using stochastic gradient descent and chose the model at the epoch with the highest F$_1$ on the development data. To initialize the embedding layers, we trained 200-dimensional word embeddings using GloVe~\cite{pennington2014} from 800 million Japanese tweets posted from March 11th, 2011 to March 11th, 2012.

\paragraph{Baselines.}
Since our methods use automatically constructed training data, we prepared two baselines that do not utilize such data.
Baseline1 regards NEs obtained by NER as emerging if they are not detected as NE on Twitter from one year to one week before the posting time of the input tweets. We set the period up to one week before to find NEs that emerge near the target day.
Baseline2 regards NEs obtained by NER as emerging if they do not exist in a KB~\cite{nakashole2013}. We regard the obtained NEs as emerging when they are not registered in Wikipedia as of the month before the posting time of the input tweets because there is a time lag to use the latest Wikipedia dump in actual settings.
To make NER robust, we use LSTM-CRF trained with a dataset combining KWDLC\footnote{\url{http://nlp.ist.i.kyoto-u.ac.jp/index.php?KWDLC}} and KNBC,\footnote{\url{http://nlp.ist.i.kyoto-u.ac.jp/kuntt/\#ga739fe2}} both of which are corpora in which NE tags are attached to noisy Web text. 

\subsection{Evaluation Procedures}
To evaluate the proposed method, we designed two evaluation procedures for emerging entities discovered from Twitter. 

\subsubsection{Precision}
To evaluate the precision of the obtained emerging entities, we applied each model to daily tweets, ranked the discovered entities using their confidence scores, and finally computed the accumulative precision for the top 500 entities. As the test sets, we randomly picked three sets of daily Japanese retweets, on March 21st, 2017 (1,695,423 tweets), June 14th, 2017 (2,041,833 tweets), and September 1st, 2017 (1,901,305 tweets) so that the seasons do not overlap. As the confidence score of Proposed (CRF) and Proposed (LSTM-CRF), we used the marginal probability obtained using the constrained forward-backward algorithm~\cite{culotta2004}. We adopted the maximum scores for the extractions when several mentions of the same entity were recognized. 
Since baselines do not provide any scores regarding the emergence of entities, we used the number of extractions of each entity normalized with the extraction number of the previous day as the confidence score. This captures the bursty feature that takes into account the appearance ratio of the previous day. 

We asked three annotators, including the first author and two student volunteers, to decide whether the outputs were accompanied by emerging contexts defined in (\S~\ref{sec:definition}) by referring to the input tweets and then adopt the majority labels to mediate the conflicts. We obtained an inter-rater agreement of 0.798 by Fleiss's Kappa~\cite{fleiss1973}, which indicates substantial agreement. This high agreement justifies the solidness of the task setting.

\subsubsection{Relative Recall and Detection Immediacy}
To evaluate the recall and detection immediacy of the obtained emerging entities, we ideally want to refer to the complete list of entities that have emerged in certain periods. However, it is unrealistic to have such a list for a diverse range of entities including long-tail emerging entities. We instead evaluated the relative recall and immediacy against a KB, by determining how many entities registered in Wikipedia could be found from the tweets and how early they were detected against their registration date in Wikipedia.

Since entities newly registered in Wikipedia include emerging and prevalent entities, we obtained the reference list of emerging entities as follows. We collected entities that appeared more than 100 times on our Twitter archive from January 1st, 2017 to June 20th, 2018, and then extracted retweets containing each entity since the first appearance. To exclude prevalent entities as much as possible, we ignored entities that appeared more than five times on our Twitter archive from March 11th, 2011 to March 11th, 2012. We obtained 13,406 entities with 9,108,612 tweets (679 tweets per entity on average) since March 12th, 2012, and then applied our method to these tweets and calculated the recall and detection immediacy of the obtained entities.

\begin{table}[!t]
\footnotesize
    \centering

        \begin{tabular}{@{\,}lr@{\,}r@{\,}r@{\,}r@{\,}}
            \toprule
            \multicolumn{1}{@{\,}l}{\textbf{daily tweets}} & \multicolumn{1}{c@{\,}}{\textbf{\textsc{head}}} & \multicolumn{1}{c@{\,}}{\textbf{\textsc{long-tail}}} & \multicolumn{1}{c@{\,}}{\textbf{\textsc{homograph}}} & \multicolumn{1}{r@{\,}}{\textbf{total}} \\
            
            \multicolumn{1}{@{\,}c}{} & \multicolumn{1}{r@{\,}}{$\mathbf{(n > 100)}$} & \multicolumn{1}{r@{\,}}{$\mathbf{(n \leq 100)}$} & \multicolumn{1}{r@{\,}}{\textbf{}} & \multicolumn{1}{r@{\,}}{} \\
            \midrule
			Mar.~21st, 2017       & 227 & 110 & 84 & 422\\
			Jun.~14th, 2017       & 214 & 106 & 77 & 397\\            
			Sep.~1st, 2017       & 261 & 110 & 66 & 437\\
			\bottomrule
        \end{tabular}
        \caption{Details of the emerging entities discovered from the daily tweets with Proposed (LSTM-CRF)}\label{table:precdetail}
\end{table}

\subsection{Results and Analysis}\label{subsec:results}
Figure 2 depicts
the cumulative precision (precision@k) for the top 500 entities discovered with each model. Proposed (LSTM-CRF) is superior to the others and mostly maintained a precision above 80\% (on average 83.2\% for top-500 entities for the three sets of daily retweets), while two Baselines remained mostly under 20\%. Proposed (LSTM-CRF) is superior to Proposed (CRF) because it models the longer contexts (entire posts) with LSTM, and can properly capture the emerging contexts detectable by seeing the entire posts.

Table~\ref{table:precdetail} lists the detected emerging entities falling under three categories to confirm whether our method could discover various types of emerging entities defined in \S~\ref{sec:definition}. \textsc{head} represents entities whose surfaces appeared over 100 times in our Twitter archive from the detection date to one year later, and \textsc{long-tail} is less than that. \textsc{homograph} represents homographic entities whose namings are already registered in Wikipedia before the detection date.
As a result, Proposed (LSTM-CRF) could discover not only entities that would be added to Wikipedia but also find many long-tail emerging entities (\textit{e.g.}, good and evil (play), Photo X Art Field (exhibition)). Their frequency is low but they are useful for companies performing social listening and local users trying to find something interest.
It also found homographic entities (\textit{e.g.}, NEVER LAND (music album), Summer of Love (musical movie)), which were not found with \textbf{Baselines}. Although it has been reported that these homographic emerging entities are difficult to find~\cite{hoffart2014,farber2016}, our method successfully discovered these entities by obtaining the emerging contexts of the entities.

\begin{table}[t]
\footnotesize
    \centering
        \begin{tabular}{@{\,}l@{\,\,\,}l@{\,}r@{\,\,\,\,}r@{\ }r@{\,}r@{\!\!\!\!\!\!\!\!\!}r@{\,}}
            \toprule
             \multicolumn{2}{@{\,}l}{\textbf{TYPE}} & \textbf{\# entities} & \multicolumn{2}{@{\,}r}{\textbf{\# found (\%)}} & \multicolumn{2}{@{\,}r}{\textbf{lead-days}} \\ 
            & \textbf{DBpedia types} & & & & 
            \multicolumn{1}{@{\,}c@{\!\!\!}}{\textbf{mean}} & \multicolumn{1}{c@{\,}}{\textbf{(median)}} \\ 
            \midrule

\multicolumn{2}{@{\,}l}{\textsc{\textbf{Person}}} & \textbf{3851} & \textbf{3238} & \textbf{(84.08\%)} & \textbf{660} & \textbf{(550)}\\ 
& Actor & 651 & 514 & (78.96\%) & 742 & (727)\\
& MusicalArtist & 624 & 463 & (74.20\%) & 740 & (649)\\
& SoccerPlayer & 477 & 429 & (89.94\%) & 769 & (759)\\
& VoiceActor & 345 & 323 & (93.62\%) & 498 & (363)\\
& AdultActor & 306 & 300 & (98.04\%) & 376 & (297)\\
& BaseballPlayer & 238 & 213 & (89.50\%) & 591 & (405)\\
& Model & 225 & 210 & (93.33\%) & 764 & (654)\\
& Person & 173 & 138 & (79.77\%) & 777 & (792)\\
& Politician & 136 & 112 & (82.35\%) & 665 & (538)\\
& Others (16 types) & 676 & 536 & (79.29\%) & 638 & (537)\\
\midrule
\multicolumn{2}{@{\,}l}{\textsc{\textbf{Creative Work}}} & \textbf{4122} & \textbf{3703} & \textbf{(89.84\%)} & \textbf{377} & \textbf{(176)}\\	
& TelevisionShow & 699 & 644 & (92.13\%) & 225 & (54)\\
& MusicSingle & 653 & 594 & (90.96\%) & 304 & (85)\\
& Film & 641 & 555 & (86.58\%) & 388 & (214)\\
& MusicAlbum & 550 & 499 & (90.73\%) & 356 & (161)\\
& Manga & 523 & 486 & (92.93\%) & 672 & (594)\\
& VideoGame & 440 & 392 & (89.09\%) & 406 & (251)\\
& RadioProgram & 228 & 203 & (89.04\%) & 261 & (38)\\
& Anime & 216 & 188 & (87.04\%) & 254 & (85)\\
& Book & 101 & 95 & (94.06\%) & 621 & (462)\\
& Others (4 types) & 71 & 47 & (66.19\%) & 700 & (515)\\
\midrule
\multicolumn{2}{@{\,}l}{\textsc{\textbf{Location}}} & \textbf{223} & \textbf{179} & \textbf{(80.27\%)} & \textbf{597} & \textbf{(385)}\\	
& Building & 89 & 73 & (82.02\%) & 497 & (291)\\
& Museum & 33 & 32 & (96.97\%) & 685 & (447)\\
& Station & 25 & 21 & (84.00\%) & 264 & (154)\\
& School & 18 & 9 & (50.00\%) & 553 & (62)\\
& Library & 13 & 13 & (100.00\%) & 995 & (1328)\\
& Park & 11 & 9 & (81.82\%) & 639 & (193)\\
& University & 7 & 6 & (85.71\%) & 904 & (996)\\
& Others (10 types) & 27 & 16 & (59.25\%) & 882 & (956)\\
\midrule
\multicolumn{2}{@{\,}l}{\textsc{\textbf{Group}}} & \textbf{240} & \textbf{152} & \textbf{(63.33\%)} & \textbf{545} & \textbf{(396)}\\	
& Company & 188 & 116 & (61.70\%) & 500 & (359)\\
& SoccerClub & 26 & 13 & (50.00\%) & 780 & (741)\\
& Organisation & 16 & 14 & (87.50\%) & 706 & (416)\\
& PoliticalParty & 10 & 9 & (90.00\%) & 552 & (471)\\
\midrule
\multicolumn{2}{@{\,}l}{\textsc{\textbf{Other}}} & \textbf{59} & \textbf{18} & \textbf{(30.51\%)} & \textbf{758} & \textbf{(977)}\\	
& Species & 53 & 14 & (26.42\%) & 825 & (1008)\\
& CelestialBody & 3 & 1 & (33.33\%) & 2 & (2)\\
& Train & 2 & 2 & (100.00\%) & 241 & (241)\\
& Aircraft & 1 & 1 & (100.00\%) & 1613 & (1613)\\
\midrule
\multicolumn{2}{@{\,}l}{\textsc{Unmapped}} & 4891 & 3562 & (72.83\%) & 690 & (615)\\
\midrule
\multicolumn{2}{@{\,}l}{\textbf{Total}}    & 13406 & 10852 & (80.95\%) & 571 & (406)\\
\bottomrule
        \end{tabular}
        \caption{Relative recall and time advantage over entity types of emerging entities detected with Proposed (LSTM-CRF)}
    \label{table:recdetail}
\end{table}

As the evaluation of relative recall, we focused on the best-performing method, \textit{i.e.,} Proposed (LSTM-CRF) and computed its relative recall over the reference list of 13,406 emerging entities.\ We detected 10,852 emerging entities (80.4\%). This is reasonably high considering that there was noise in the reference list such as a concept name that was defined after it became prevalent (\textit{e.g.}, Virtual Youtuber) and periodic entities (\textit{e.g.}, Tokyo prefectural election, 2017).

Table~\ref{table:recdetail} shows the distribution of the types of the 13,406 entities obtained by the DBpedia mappings, detection ratio, and lead-time against the Wikipedia registration time for each type. 
For \textsc{Person}, \textsc{Creative Work}, and \textsc{Location} types, our model found on average more than 80\% of entities. 
On the other hand, for \textsc{Group} and \textsc{Other} types, detection rates dropped remarkably. We found that some of those entities do not appear in emerging contexts at all within our Twitter archive. Since our method utilizes such emergence signals as the clue, it is difficult to discover entities appearing without emerging contexts. This is the current limitation of our method. Note that the 13,406 entities used in this evaluation included some prevalent entities (\textit{e.g.,} local company) that might also affect the performances.

We next evaluated detection immediacy.\ We found that 92.4\% of the discovered entities (10,030 out of 10,852) were detected earlier than their registration in Wikipedia\@. We then investigated the remaining 822 (7.6\%) entities and found that they were mostly periodic events such as Olympics and election, or incorrectly included prevalent entities. 
The mean (and median) lead days of the first day when Proposed (LSTM-CRF) detected each entity against their registration date were 571 (and 406) days, which supports the detection immediacy of our method.\ Compared to \textsc{Creative Work} types of entities, our method detected \textsc{Person} and \textsc{Location} types of entities earlier than their registration in Wikipedia\@, which means those entity types take longer to be notable enough to be registered in Wikipedia~\cite{graus2018}. 

Overall, these results reconfirm that microblogs are useful sources for finding emerging entities and our method can detect such entities at the early-stage of their appearance. It also implies that relying on Wikipedia for source of entities misses valuable information on emerging entities.

\section{Conclusions}
We introduced a novel task of discovering emerging entities in microblogs (\S~\ref{sec:intro}, \ref{sec:definition}). We pointed out the problems of related tasks (\S~\ref{sec:related}) and proposed an effective method for discovering emerging entities in microblogs by exploiting the contexts of those entities using time-sensitive distant supervision (\S~\ref{sec:proposal}). Experimental results demonstrated that our method performed accurately and showed that emerging entities, including homographic and long-tail ones, can be effectively and instantly discovered by obtaining emerging contexts (\S~\ref{sec:experiment}).

We plan to carry out semantic typing of emerging entities by exploiting emerging contexts, which are likely to include enough information for the public to understand the entities. 

\section*{Acknowledgments}
This work was supported by JSPS KAKENHI Grant Number 18J22830. 
The authors thank the volunteers and the anonymous reviewers for their hard work.

\newpage
\bibliographystyle{named}
\bibliography{www19}

\begin{thebibliography}{}

\bibitem[\protect\citeauthoryear{Akbik \bgroup \em et al.\egroup
  }{2019}]{akbik:2019}
Alan Akbik, Tanja Bergmann, and Roland Vollgraf.
\newblock Pooled contextualized embeddings for named entity recognition.
\newblock In {\em Proceedings of the 2019 Annual Conference of the North
  American Chapter of the Association for Computational Linguistics: Human
  Language Technologies (NAACL-HLT)}, pages 724--728, 2019.

\bibitem[\protect\citeauthoryear{Brambilla \bgroup \em et al.\egroup
  }{2017}]{brambilla2017}
Marco Brambilla, Stefano Ceri, Emanuele Della~Valle, Riccardo Volonterio, and
  Felix~Xavier Acero~Salazar.
\newblock Extracting emerging knowledge from social media.
\newblock In {\em Proceedings of the 26th International Conference on World
  Wide Web (WWW)}, pages 795--804, 2017.

\bibitem[\protect\citeauthoryear{Brown \bgroup \em et al.\egroup
  }{1992}]{brown1992}
Peter~F Brown, Peter~V Desouza, Robert~L Mercer, Vincent J~Della Pietra, and
  Jenifer~C Lai.
\newblock Class-based $n$-gram models of natural language.
\newblock {\em Computational linguistics}, 18(4):467--479, 1992.

\bibitem[\protect\citeauthoryear{Culotta and McCallum}{2004}]{culotta2004}
Aron Culotta and Andrew McCallum.
\newblock Confidence estimation for information extraction.
\newblock In {\em Proceedings of the 5th Annual Conference of the North
  American Chapter of the Association for Computational Linguistics: Human
  Language Technologies (NAACL-HLT)}, pages 109--112, 2004.

\bibitem[\protect\citeauthoryear{Derczynski \bgroup \em et al.\egroup
  }{2017}]{derczynski2017}
Leon Derczynski, Eric Nichols, Marieke van Erp, and Nut Limsopatham.
\newblock Results of the {WNUT2017} shared task on novel and emerging entity
  recognition.
\newblock In {\em Proceedings of the 3rd Workshop on Noisy User-generated Text
  (WNUT)}, pages 140--147, 2017.

\bibitem[\protect\citeauthoryear{F{\"a}rber \bgroup \em et al.\egroup
  }{2016}]{farber2016}
Michael F{\"a}rber, Achim Rettinger, and Boulos Asmar.
\newblock On emerging entity detection.
\newblock In {\em Proceedings of the 20th International Conference on Knowledge
  Engineering and Knowledge Management (EKAW)}, pages 223--238, 2016.

\bibitem[\protect\citeauthoryear{Fleiss and Cohen}{1973}]{fleiss1973}
Joseph~L Fleiss and Jacob Cohen.
\newblock The equivalence of weighted kappa and the intraclass correlation
  coefficient as measures of reliability.
\newblock {\em Educational and psychological measurement}, 33(3):613--619,
  1973.

\bibitem[\protect\citeauthoryear{Graus \bgroup \em et al.\egroup
  }{2018}]{graus2018}
David Graus, Daan Odijk, and Maarten de~Rijke.
\newblock The birth of collective memories: Analyzing emerging entities in text
  streams.
\newblock {\em Journal of the Association for Information Science and
  Technology}, 69(6):773--786, 2018.

\bibitem[\protect\citeauthoryear{Hoffart \bgroup \em et al.\egroup
  }{2014}]{hoffart2014}
Johannes Hoffart, Yasemin Altun, and Gerhard Weikum.
\newblock Discovering emerging entities with ambiguous names.
\newblock In {\em Proceedings of the 23rd International Conference on World
  Wide Web (WWW)}, pages 385--396, 2014.

\bibitem[\protect\citeauthoryear{Lafferty \bgroup \em et al.\egroup
  }{2001}]{lafferty2001}
John~D. Lafferty, Andrew McCallum, and Fernando C.~N. Pereira.
\newblock Conditional random fields: Probabilistic models for segmenting and
  labeling sequence data.
\newblock In {\em Proceedings of the 18th International Conference on Machine
  Learning (ICML)}, pages 282--289, 2001.

\bibitem[\protect\citeauthoryear{Lample \bgroup \em et al.\egroup
  }{2016}]{lample2016}
Guillaume Lample, Miguel Ballesteros, Sandeep Subramanian, Kazuya Kawakami, and
  Chris Dyer.
\newblock Neural architectures for named entity recognition.
\newblock In {\em Proceedings of the 15th Annual Conference of the North
  American Chapter of the Association for Computational Linguistics: Human
  Language Technologies (NAACL-HLT)}, pages 260--270, 2016.

\bibitem[\protect\citeauthoryear{McCallum}{2002}]{McCallumMALLET}
Andrew~Kachites McCallum.
\newblock Mallet: A machine learning for language toolkit.
\newblock http://mallet.cs.umass.edu, 2002.

\bibitem[\protect\citeauthoryear{Miller \bgroup \em et al.\egroup
  }{2004}]{miller2004}
Scott Miller, Jethran Guinness, and Alex Zamanian.
\newblock Name tagging with word clusters and discriminative training.
\newblock In {\em Proceedings of the 5th Annual Conference of the North
  American Chapter of the Association for Computational Linguistics: Human
  Language Technologies (NAACL-HLT)}, pages 337--342, 2004.

\bibitem[\protect\citeauthoryear{Mintz \bgroup \em et al.\egroup
  }{2009}]{mintz2009}
Mike Mintz, Steven Bills, Rion Snow, and Dan Jurafsky.
\newblock Distant supervision for relation extraction without labeled data.
\newblock In {\em Proceedings of the Joint Conference of the 47th Annual
  Meeting of the Association for Computational Linguistics and the 4th
  International Joint Conference on Natural Language Processing (ACL-IJCNLP)},
  pages 1003--1011, 2009.

\bibitem[\protect\citeauthoryear{Nakashole \bgroup \em et al.\egroup
  }{2013}]{nakashole2013}
Ndapandula Nakashole, Tomasz Tylenda, and Gerhard Weikum.
\newblock Fine-grained semantic typing of emerging entities.
\newblock In {\em Proceedings of the 51st Annual Meeting of the Association for
  Computational Linguistics (ACL)}, pages 1488--1497, 2013.

\bibitem[\protect\citeauthoryear{Pennington \bgroup \em et al.\egroup
  }{2014}]{pennington2014}
Jeffrey Pennington, Richard Socher, and Christopher Manning.
\newblock Glove: Global vectors for word representation.
\newblock In {\em Proceedings of the 19th Conference on Empirical Methods in
  Natural Language Processing (EMNLP)}, pages 1532--1543, 2014.

\bibitem[\protect\citeauthoryear{Ratinov and Roth}{2009}]{ratinov2009}
Lev Ratinov and Dan Roth.
\newblock Design challenges and misconceptions in named entity recognition.
\newblock In {\em Proceedings of the 13th Conference on Computational Natural
  Language Learning (CoNLL)}, pages 147--155, 2009.

\bibitem[\protect\citeauthoryear{Wu \bgroup \em et al.\egroup }{2016}]{wu2016}
Zhaohui Wu, Yang Song, and C~Lee Giles.
\newblock Exploring multiple feature spaces for novel entity discovery.
\newblock In {\em Proceedings of the 30th AAAI Conference on Artificial
  Intelligence (AAAI)}, pages 3073--3079, 2016.

\bibitem[\protect\citeauthoryear{Yang \bgroup \em et al.\egroup
  }{2018}]{yang2018}
Jie Yang, Shuailong Liang, and Yue Zhang.
\newblock Design challenges and misconceptions in neural sequence labeling.
\newblock In {\em Proceedings of the 27th International Conference on
  Computational Linguistics (COLING)}, pages 3879--3889, 2018.

\end{thebibliography}

\end{document}